\documentclass[letterpaper, 10 pt, conference]{ieeeconf}  %

\IEEEoverridecommandlockouts                              %

\usepackage{hyperref} 
\usepackage{mathrsfs}
\usepackage{amsfonts}
\usepackage{amsmath}
\usepackage[flushleft]{threeparttable}
\usepackage{tablefootnote}
\usepackage{multirow}
\usepackage{graphicx}
\usepackage{subfigure}
\let\labelindent\relax
\usepackage{enumitem}
\usepackage{hanging}
\usepackage{color}
\usepackage{tikz}
\usetikzlibrary{calc,arrows,decorations.markings}
\usepackage{balance}
\usepackage{cite}

\graphicspath{{image}}

\usepackage{booktabs}  %
\usepackage{colortbl}  %

\usepackage{cuted}
\usepackage{capt-of}

\title{\LARGE \bf Parallel Inversion of Neural Radiance Fields for Robust Pose Estimation}

\author{Yunzhi Lin$^{1,2}$, Thomas M{\"u}ller$^{1}$, Jonathan Tremblay$^{1}$, Bowen Wen$^{1}$, Stephen Tyree$^{1}$, \\
Alex Evans$^{1}$, Patricio A.  Vela$^{2}$, 
Stan Birchfield$^{1}$ 
\\ $^{1}$NVIDIA: {\tt\small \{tmueller, jtremblay, 
bowenw, styree, alexe, sbirchfield\}@nvidia.com}
\\ $^{2}$Georgia Institute of Technology: {\tt\small \{yunzhi.lin, pvela\}@gatech.edu}
\thanks{Work was completed while the first author was an intern at NVIDIA.}%
}

\begin{document}

\twocolumn[{%
\renewcommand\twocolumn[1][]{#1}%
\maketitle

\begin{center}
    \vspace*{-5mm}
    \centering
    \begin{tabular}{ccc}
    {\includegraphics[width=0.98\textwidth,clip=true,trim=0in 0in 0in 0in]{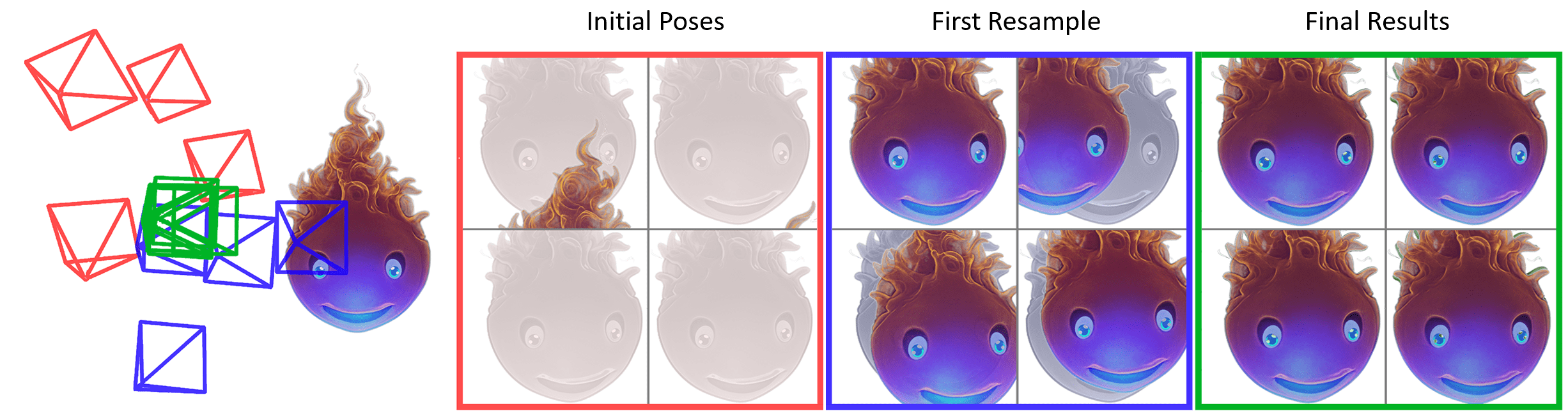}} 
    \vspace*{-2mm}
    \end{tabular}
    \captionof{figure}{\textit{Our NeRF-based parallelized optimization method estimates the camera pose from a monocular RGB image of a novel object.} The optimization iteratively updates a set of pose estimates in parallel by backpropagating discrepancies between the observed and rendered image. 
    {\sc Left:} To simplify the display, we show four camera hypotheses at three iterations:  initial (red), after the first resample (blue), and final (green).
    {\sc Right:} Corresponding renderings of estimated poses (in color) overlaid on the observed image (grayscale).
    \label{fig:abstract_demo}}
\end{center}%
}]

\begin{abstract}

We present a parallelized optimization method based on fast Neural Radiance Fields (NeRF) for estimating 6-DoF pose of a camera with respect to an object or scene.
Given a single observed RGB image of the target, we can predict the translation and rotation of the camera by minimizing the residual between pixels rendered from a fast NeRF model and pixels in the observed image.
We integrate a momentum-based camera extrinsic optimization procedure into Instant Neural Graphics Primitives, a recent exceptionally fast NeRF implementation.
By introducing parallel Monte Carlo sampling into the pose estimation task, our method overcomes local minima and improves efficiency in a more extensive search space.
We also show the importance of adopting a more robust pixel-based loss function to reduce error.
Experiments demonstrate that our method can achieve improved generalization and robustness on both synthetic and real-world benchmarks.\makeatletter{\renewcommand*{\@makefnmark}{}\footnotetext{Project page: \url{https://pnerfp.github.io}}}

\end{abstract}

\section{Introduction}

6-DoF pose estimation---predicting the 3D position and orientation of a camera with respect to an object or scene---is a fundamental step for many tasks,  including some in robot manipulation and augmented reality. While RGB-D or point cloud-based methods~\cite{wang2019densefusion, he2020pvn3d, he2021ffb6d,wen2021bundletrack} have received much attention,  monocular RGB-only approaches~\cite{tremblay2018deep, lin2022single} have great potential for wider applicability and for handling certain material properties that are difficult for depth sensors---such as transparent or dark surfaces.

Much research in this area has focused on instance-level object pose estimation
\cite{xiang2017posecnn, sundermeyer2018implicit, peng2019pvnet, wang2021gdr, haugaard2022surfemb}. These methods assume that a textured 3D model of the target object is available for training. Such methods have achieved success under different scenarios but suffer from a lack of scalability. Research beyond this limitation considers category-level object pose estimation ~\cite{manhardt2020cps++, hou2020mobilepose,ahmadyan2021objectron, lin2022single}. These methods scale better for real-world applications, since a single trained model works for a variety of object instances within a known category. Nevertheless, the effort needed to define and train a model for each category remains a limitation.   
For more widespread generalizability, it is important to be able to easily estimate poses for arbitrary objects.

The emergence of Neural Radiance Fields (NeRF)~\cite{mildenhall2020nerf} has the potential to facilitate novel object pose estimation. 
NeRF and its variants learn generative models of objects from pose-annotated image collections, capturing complex 3D structure and high-fidelity surface details. Recently, iNeRF~\cite{yen2021inerf} has been proposed as an  analysis-by-synthesis approach for pose estimation built on the concept of inverting a NeRF model. 
Inspired by iNeRF's success, this paper further explores the idea of pose estimation via neural radiance field inversion.

A drawback of NeRF is its computational overhead which impacts execution time.
To overcome this limitation, we leverage our fast version of NeRF, known as Instant Neural Graphics Primitives (Instant NGP)~\cite{mueller2022instant}. Using Instant NGP model inversion provides significant speedups over NeRF.
The structure of Instant NGP admits parallel optimization, which is leveraged to overcome issues with local minima and thereby achieve greater robustness than possible with iNeRF.
Similar to iNeRF, our pose estimation requires three inputs: 
a single RGB image with the target,  
an initial rough pose estimate of the target, and 
an instant NGP model trained from multiple views of the target.

Considering that a single camera pose is vulnerable to local minima during optimization iterations, we leverage parallelized Monte Carlo sampling. At adaptive intervals, camera pose hypotheses are 
re-sampled around the hypotheses with the lowest loss. This design alleviates the issue of convergence to local minima and improves efficiency of search over a more extensive search space. 

The gradients of pixel residuals calculated between the rendered model and the target view are backpropagated to generate camera pose updates. Unlike iNeRF where a subsample of a new image is rendered at each iteration,%
we enable hundreds of thousands of rays to work independently in parallel to accumulate gradient descent updates per camera pose hypothesis. This design dramatically improves the efficiency. 
Furthermore,  we investigate different pixel-based loss functions to identify which approach to quantifying the visual difference between the rendered model and the observed target image best informs the camera pose updates.
As shown in the ablation study, the mean absolute percentage error (MAPE)~\cite{myttenaere2015esann:mape} loss exhibits better robustness to disturbances.

In summary, this work makes the following contributions:
\begin{itemize}[leftmargin=*]
    \item A parallelized, momentum-based optimization method using NeRF models is proposed to estimate 6-DoF poses from monocular RGB input. 
    The object-specific NeRF model does not require pre-training on large datasets. 
    \item Parallelized Monte Carlo sampling is introduced into the pose estimation task, and we show the importance of pixel-based loss function selection for robustness. 
    \item Quantitative demonstration through synthetic and real-world benchmarks that the proposed method has improved generalization and robustness.
\end{itemize}

\section{Related Works}

{\bf Neural 3D Scene Representation.} 
Recent works~\cite{mescheder2019occupancy, park2019deepsdf, sitzmann2020metasdf} have investigated representing 3D scenes implicitly with neural networks, where coordinates are sampled and fed into a neural network to produce physical field values across space and time~\cite{xie2022neural}.
NeRF~\cite{mildenhall2020nerf} is a milestone approach demonstrating that neural scene representations have the capabilities to synthesize photo-realistic views.
Since then, significant effort has been put into pushing the boundaries of NeRF. Follow-up works have focused on speeding up the training and inference processes~\cite{yu2021plenoctrees, reiser2021kilonerf, mueller2022instant}, adding support for relighting~\cite{zhang2021physg}, relaxing the requirement of known camera poses~\cite{lin2021barf, wang2021nerfmm}, reducing the number of training images~\cite{chen2021mvsnerf}, extending to dynamic scenes~\cite{li2021neural}, and so on.
NeRF also opens up opportunities in the robotics community. Researchers have proposed to use it to represent scenes for visuomotor control~\cite{li20223d}, reconstruct transparent objects~\cite{ichnowski2022dex}, generate training data for pose estimators~\cite{li2022nerf} or dense object descriptors~\cite{yen2022nerf}, and model 3D object categories~\cite{xie2021fig}.
In this work, we aim to follow in their footsteps by applying NeRF directly to the 6-DoF pose estimation task.

{\bf Generalizable 6-DoF Pose Estimation.} 
Generalizable 6-DoF pose estimation---not limited to any specific target or category---from RGB images has been a long-standing problem in the community. Existing methods tend to share a similar pipeline of two phases:  1) model registration and 2) pose estimation.

Traditional methods~\cite{pauwels2015simtrack, trinh2018modular, pitteri20203d,wen2020robust} first build a 3D CAD model via commercial scanners or dense 3D reconstruction techniques~\cite{schonberger2016pixelwise, alicevision2021}. They resolve the pose by finding 2D-3D correspondences (via hand-designed features like SIFT~\cite{lowe2004distinctive} or ORB~\cite{rosten2006machine}) between the input RGB image and the registered model. However, creating high quality 3D models is not easy, %
and finding correspondence across a large database (renderings from different viewpoints) can be time-consuming~\cite{pauwels2015simtrack}. 
More recently, several attempts have been made to revisit the object-agnostic pose estimation problem with deep learning. The presumption is that a deep network pretrained on a large dataset can generalize to find correspondence between the query image and the registered model for novel objects.  
OnePose~\cite{sun2022onepose}, inspired by visual localization research, proposes to use a graph attention network to aggregate 2D features from different views during the registration phase of structure-from-motion~\cite{schonberger2016structure}. Then the aggregated 3D descriptor is matched with 2D features from the query view to solve the P$n$P problem~\cite{lepetit2009epnp}. 
Similarly, OSOP~\cite{shugurov2022osop} explores solving the P$n$P problem with a dense correspondence between the query image and the coordinate map from a pre-built 3D CAD model. On the other hand, Gen6D~\cite{liu2022gen6d} only needs to register the model with a set of posed images. Following the iterative template matching idea~\cite{li2018deepim,wen2020se}, its network takes as input several neighboring registered images closest to the predicted pose and repeatedly refines the result.

While data-driven approaches rely on the generalization of a large training dataset (usually composed of both synthetic \& real-world data)~\cite{liu2022gen6d}, iNeRF~\cite{yen2021inerf} is an optimization on-the-fly approach free of pretraining. Each new object is first registered by a NeRF model~\cite{mildenhall2020nerf}, after which iNeRF can optimize the camera pose on the synthesized photo-realistic renderings from NeRF. Although iNeRF's idea seems promising, there still remain several challenges. 
The first is the expensive training cost of a NeRF model, which may take hours for just one target. 
Additionally, iNeRF's pose update strategy is inefficient, as the accumulation and backpropagation of the loss gradient is performed until a subsample of a new image is rendered.
Moreover, the optimization process of a single pose hypothesis is easily trapped in local minima due to outliers. To deal with the aforementioned issues, we propose a more efficient and robust approach leveraging the recent success of Instant NGP~\cite{mueller2022instant}. We re-formulate the camera pose representation as the Cartesian product ${\mathrm{SO}(3) \times \mathrm{T}(3)}$ and integrate the optimization process into the structure of Instant NGP. We also adopt parallelized Monte Carlo sampling to improve robustness to local minima.
Loc-NeRF~\cite{maggio2022arx:locnerf} is another concurrent work using Monte Carlo sampling to improve iNeRF.

\section{Preliminaries}

{\bf NeRF.}
Given a collection of $N$ RGB images $\left\{I_i\right\}_{i=1}^N, I_i \in$ $[0,1]^{H \times W \times 3}$ with known camera poses $\left\{T_i\right\}_{i=1}^N$, NeRF~\cite{mildenhall2020nerf} learns to represent a scene as 5D neural radiance fields (spatial location and viewing direction). It can synthesize novel views by querying 5D coordinates along the camera rays and use classic volume rendering techniques to project the output colors and densities into an image. 

{\bf Instant NGP.}
To further reduce the training and inference cost of the vanilla NeRF~\cite{mildenhall2020nerf}, Instant NGP~\cite{mueller2022instant} proposes to adopt a small neural network augmented by a multi-resolution hash table of trainable feature vectors. This structure allows the network to disambiguate hash collisions, making it easy to parallelize on GPUs. The method achieves a combined speedup of several orders of magnitude, allowing its use in time-constrained settings like online training and inference.

{\bf iNeRF's Formulation.}
Assuming the target scene has been trained with a NeRF model parameterized with weight $\Theta$ and the camera intrinsics are
known, iNeRF~\cite{yen2021inerf} aims to recover the camera pose $T \in SE(3)$ of an observed image $I$ given the weight $\Theta$:
\begin{equation}
\hat{T}=\underset{T \in \operatorname{SE}(3)}{\operatorname{argmin}}\, \mathcal{L}(T \mid I, \Theta)
\end{equation}
with $\mathcal{L}$ the loss between the NeRF rendering and the observed image. It uses L2 loss in practice.
In the optimization process, iNeRF fixes the NeRF's weight $\Theta$ and iteratively updates $T$ to minimize $\mathcal{L}$.

\section{Approach}

\begin{figure}[t]
\centering
\hspace*{-3.5mm}\includegraphics[width=0.8\columnwidth]{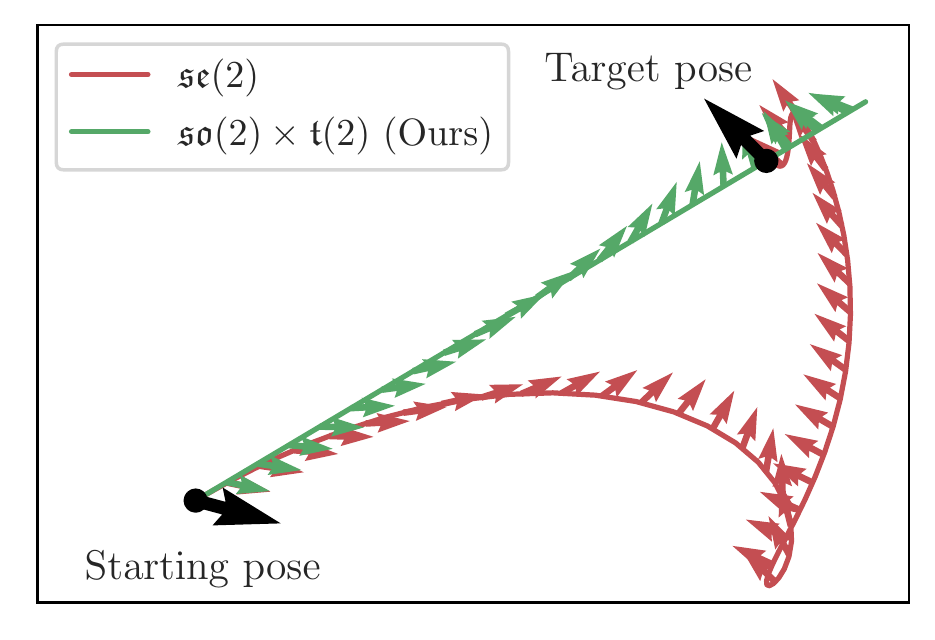}%
\vspace{-0.15in}%
\caption{\label{fig:so2-cross-t2-vs-se2}{\sc Green:} Our decomposition of gradients and their momentum into the rotational Lie algebra $\mathfrak{so}(2)$ and the translational Lie algebra $\mathfrak{t}(2)$ yields a straight path from the starting pose to the target pose using gradient descent with momentum, converging in 69 steps.
Momentum causes our optimization to overshoot and snap back to the target, all along the straight line.
{\sc Red:} Using the special Euclidean Lie algebra $\mathfrak{se}(2)$~\cite{yen2021inerf, lin2021barf} leads to a suboptimal curved path due to coupling between translation and rotation, requiring 85 steps to converge.
}
  \vspace*{-2 ex}
\end{figure}
\subsection{Momentum-based Camera Extrinsics Optimization}

For this work, we modified the Instant NGP~\cite{mueller2022instant} camera pose and gradient representations from their standard use in NeRF. Critically, this permitted the dynamics of the gradient updates to incorporate momentum-based approaches for enhanced optimization. The section details those changes.

\paragraph{Camera Pose Representation}
Camera poses consist of a translation component (position) as well as a rotation component (orientation) and are often modeled by the special Euclidean group in 3D, $SE(3)$.
The goal of extrinsics optimization in NeRF~\cite{yen2021inerf, lin2021barf} is to find those camera poses that minimize the image-space loss by gradient descent.
Gradient updates are computed in the special Euclidean Lie algebra $\mathfrak{se}(3)$, then applied to generate a camera pose update combining rotation and translation.
However, using a native $SE(3)/\mathfrak{se}{3}$ representation has a disadvantage: a camera pose update's center of rotation is not at the camera origin, but on the screw axis, which couples camera position and orientation. This coupling can lead to nonintuitive optimization trajectories when momentum is involved.
To decouple the translation and rotation updates, we model camera pose as the Cartesian product ${\mathrm{SO}(3) \times \mathrm{T}(3)}$ (and likewise the respective Lie algebra, ${\mathfrak{so}(3) \times \mathfrak{t}(3)}$), which employs an additive structure on $\mathrm{T}(3)$ and a product structure on $\mathrm{SO}(3)$. 
Momentum in this representation moves in straight lines and rotates along geodesic paths over the surface of the sphere, converging in fewer steps.

Fig.~\ref{fig:so2-cross-t2-vs-se2} shows a 2D example, where we use the momentum-based Adam optimizer~\cite{kingma2014adam} to minimize the mean squared error (MSE) between the $3\times3$ matrices describing the target pose and the current pose.
The gradient of the MSE always points along the straight line to the target in 2D space, yet the trajectory through $\mathfrak{se}(2)$ (shown in red) deviates from the straight line, because of the aforementioned coupling.
Note that the trajectory \emph{starts out} straight---gradients without momentum always point straight to the target, regardless of representation---begins to curve once momentum builds up, which becomes increasingly inaccurate as $SE(2)$ is traversed. The trajectory eventually converges as the momentum is corrected over the course of many optimization steps.

\paragraph{Momentum-Based Optimization}
Contemporary momentum-based optimization has empirically demonstrated more effective convergence properties over standard gradient-based approaches, especially when combined with an adaptive update law. For implementation of the optimization update laws, the Adam optimizer~\cite{kingma2014adam} with first and second moments is applied independently on the two subspaces of our representation, $\mathrm{SO}(3)$ and $\mathrm{T}(3)$. 
Crucially, the momentum-based updates are cheaply and stably computed from the NeRF implementation.
In NeRF, each pixel corresponds to a ray with origin $\mathrm{o}$ and direction $\mathrm{d}$, along which the model is evaluated at $K$ positions, $\mathrm{p}_i = \mathrm{o} + t_i \cdot \mathrm{d}$, based on moving distances $t_i$ along the ray.
In our decomposition, the ray origin $\mathrm{o}$ is the same as the camera origin, and thus corresponds to $\mathrm{T}(3)$.
Its gradient in $\mathfrak{t}(3)$ with respect to the image-space loss $\mathcal{L}$ is
\begin{align}\label{eq:trans-grad}
\frac{\partial\,\mathcal{L}}{\partial\,\mathrm{o}} = \sum_{i=1}^K \frac{\partial\,\mathcal{L}}{\partial\,\mathrm{p}_i} \,,
\end{align}
averaged over all pixels of the image, where ${\partial\,\mathcal{L} / \partial\,\mathrm{p}_i}$ is obtained from standard backpropagation through the NeRF algorithm.
Similarly, the ray direction $\mathrm{d}$ is rigidly coupled to the camera orientation, whose gradient in $\mathfrak{so}(3)$ is stably computed as
\begin{align}\label{eq:rot-grad}
\mathrm{\tau}(\mathrm{d}) = \sum_{i=1}^K \, t_i \cdot \left(\mathrm{d} \times \frac{\partial\,\mathcal{L}}{\partial\,\mathrm{p}_i}\right) \,,
\end{align}
also averaged over all pixels of the image. The direction of the vector $\mathrm{\tau}(\mathrm{d})$ is the rotation axis of the orientation gradient and its length the magnitude.
Momentum-based optimizers, such as Adam~\cite{kingma2014adam}, must maintain their rotational moments as vectors computed from $\mathrm{\tau}(\mathrm{d})$ and the current orientation as a ${3\times3}$ matrix whose updates are rotations induced by these vector moments.

A physical interpretation based on rigid-body mechanics is as follows:  Imagine each ray-derived point as being physically attached to the camera, and the image-based loss function gradient as a force applied to that point.  Then  the influence is a translational force (Eq.~\eqref{eq:trans-grad}) and a torque (Eq.~\eqref{eq:rot-grad}), both acting on the camera. 
Hence, application of this decomposition to the Adam optimizer~\cite{kingma2014adam} turns Adam's first moment into physical momentum for cameras being ``pushed around'' by the gradients acting as forces---although Adam's second moment and exponential decay do not have straightforward physical analogues.

\begin{table*}[t] 
 \vspace*{0.06in}
  \caption{6-DoF pose estimation results on the NeRF synthetic and LLFF datasets. 
  \label{tab:synthetic_and_llff}}  
  \centering 
  \begin{tabular}{ccccccccc>{\columncolor[gray]{0.9}[1pt]}ccccc>{\columncolor[gray]{0.9}[1pt]}c}
\toprule
\addlinespace
& \multicolumn{9}{c}{(a) NeRF Synthetic} & \multicolumn{5}{c}{(b) LLFF} \\
\addlinespace
Method                & Chair & Drums & Ficus & Hotdog & Lego & Materials & Mic  & Ship & \textbf{Mean}  & Fern & Fortress & Horns & Room & \textbf{Mean}  \\
\cmidrule(r){2-10} \cmidrule(r){11-15}
& \multicolumn{9}{c}{Rotation error $<5^{\circ}$ ($\uparrow$)} 
& \multicolumn{5}{c}{Rotation error $<5^{\circ}$ ($\uparrow$)} \\
\addlinespace

iNeRF~\cite{yen2021inerf}           & 0.44  & 0.48  & 0.64  & 0.28   & 0.68 & 0.76      & 0.24 & 0.40  & 0.49 & 0.60 & 0.72     & 0.64   & 0.52 & 0.62 \\
Ours (single)   & 0.88  & 0.52  & 0.72  & 0.80    & 0.80  & 0.84      & 0.48 & 0.76 & 0.73 & 0.88 & 0.84     & 0.80   & 0.84 & 0.84 \\
Ours (multiple) & {\bf 1.00}     &  {\bf 1.00}      &  {\bf 1.00}      &  {\bf 1.00}       &  {\bf 1.00}     &  {\bf 1.00}          &  {\bf 1.00}     & {\bf 0.84} & {\bf 0.98} &  {\bf 1.00}    &      {\bf 1.00}      &      {\bf 1.00}   &     {\bf 1.00}   &   {\bf 1.00}   \\
\vspace{-7pt} \\
\cmidrule(r){2-10} \cmidrule(r){11-15}
& \multicolumn{9}{c}{Translation error $<0.05$ units ($\uparrow$)} 
& \multicolumn{5}{c}{Translation error $<0.05$ units ($\uparrow$)} \\
\addlinespace

iNeRF~\cite{yen2021inerf}           & 0.44  & 0.52  & 0.64  & 0.32   & 0.76 & 0.56      & 0.20  & 0.40  & 0.48 & 0.56 & 0.72     & 0.64  & 0.48 & 0.60  \\
Ours (single)   & 0.88  & 0.52  & 0.72  & 0.76   & 0.80  & 0.56      & 0.28 & {\bf 0.40}  & 0.62 & 0.76 & 0.76     & 0.76  & 0.84 & 0.78 \\
Ours (multiple) &  {\bf 1.00}      &  {\bf 1.00}      &  {\bf 1.00}      & {\bf 0.96}   &  {\bf 1.00}     &  {\bf 1.00}          & {\bf 0.84} & 0.32 & {\bf 0.89} &   {\bf 1.00}     &     {\bf 1.00}       &     {\bf 1.00}    &   {\bf 1.00}     &   {\bf 1.00}  \\
\bottomrule
  \end{tabular}
  \vspace*{-2 ex}
\end{table*} 

\begin{figure}[t]
  \centering 
    \includegraphics[width=\columnwidth,clip=true,trim=0in 0in 0in 0in]{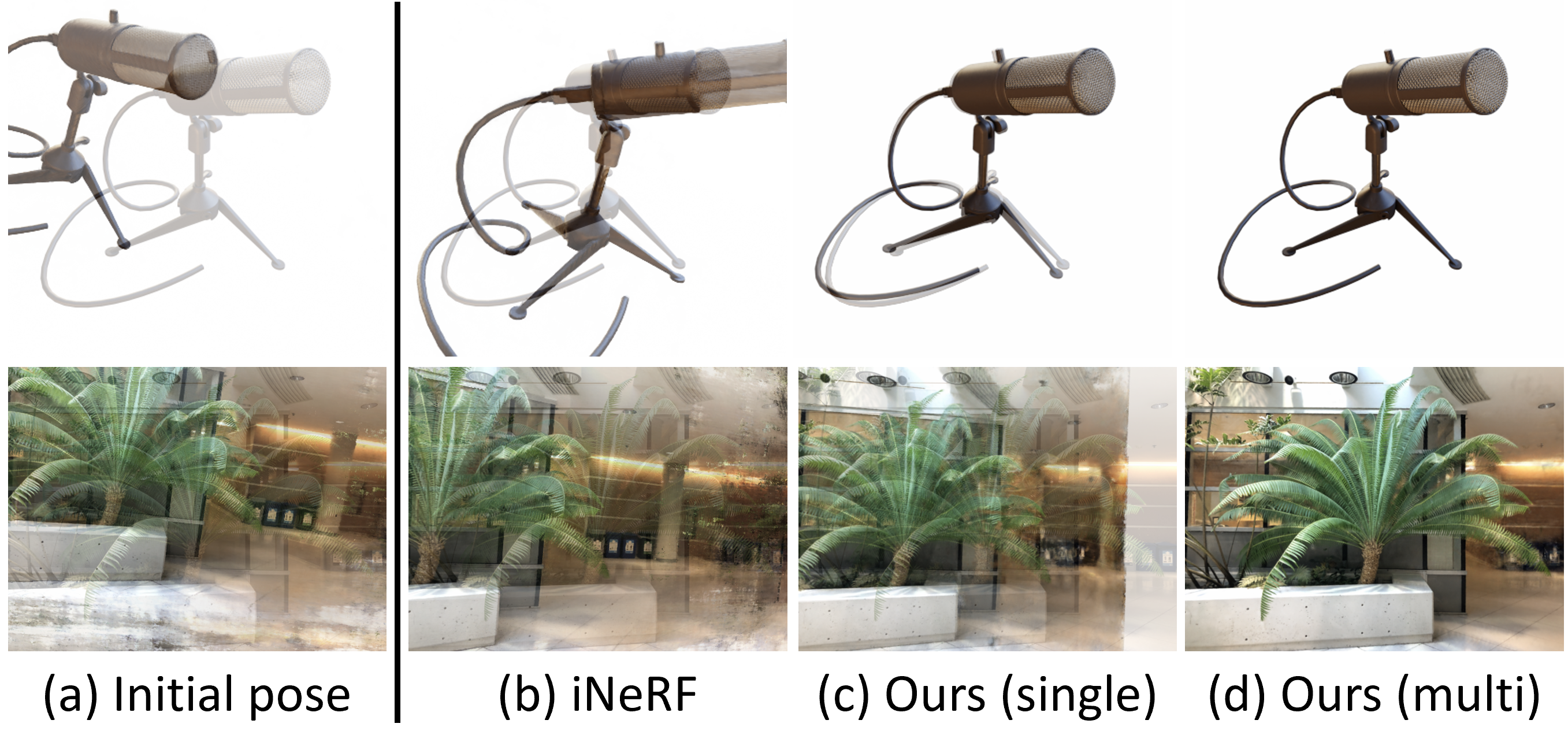}
  \caption{A qualitative comparison between iNeRF~\cite{yen2021inerf} and our proposed method on both synthetic dataset~\cite{mildenhall2020nerf} and real-world dataset~\cite{mildenhall2019local}, where the rendered model under the estimated pose (color) is blended with the observed image (white). We show (a) the initial pose for all the methods while (b)-(d) present the final optimized poses for different approaches.
  \label{fig:result_comp}} 
  \vspace*{-15 px}
\end{figure}

\subsection{Parallelized Monte Carlo Sampling}

As the loss function we optimize is non-convex over the 6-DoF space~\cite{yen2021inerf}, it is easy for a single camera pose hypothesis to be trapped in local minima. 
Thanks to the computing capacity of Instant NGP~\cite{mueller2022instant}, we are able to start the optimization from multiple hypotheses simultaneously. 
However, a simple multi-start idea is inefficient, especially in a large search space, where many hypotheses would be way off during the optimization process. As a result, they cannot contribute to the final optimization but still occupy a lot of computing resources.

We draw inspiration from the particle filtering framework~\cite{douc2005comparison, choi2012robust, deng2021poserbpf} and propose a simple and effective pose hypothesis update strategy to handle this problem.
We divide the optimization process into two phases, 1) free exploration and 2) resampling update.  
In the first phase, we generate $P_N$ camera pose hypotheses around the start pose, with translation and rotation offsets uniformly sampled in the Euclidean space and $\mathrm{SO}(3)$, respectively. The camera pose hypotheses are optimized independently by~\cite{mueller2022instant} for $s_1$ steps. It is expected that at least some of these hypotheses will move toward the ground truth. Next, we move to the second phase, where the losses of all hypotheses are measured and taken as a reference for the sampling weight. We keep the first $r$ ratio of the hypotheses with the lowest losses and resample the remaining hypotheses around them with small offsets. The hypotheses are optimized independently again for $s_2$ steps.  The second phase is repeated $S$ times,  reducing $r$ each round.

\subsection{Pixel-based RGB Loss \label{Approach_loss}}

One of the biggest challenges for analysis-by-synthesis pose estimation methods is that the registered model will have a different appearance than the target view, even when rendered in same pose. This issue is the result of changes in, e.g., lighting condition, occlusion, and environmental noise. While previous work such as iNeRF~\cite{yen2021inerf} follows a common practice to employ an L2 loss, we investigate additional losses to measure the difference between rendered and observed pixels. These loss functions vary in their treatment of visual errors as well as in their convergence properties, which in turn affect the optimization process.

Since our basic NeRF model (Instant NGP~\cite{mueller2022instant}) treats individual sampling rays independently, we focus exclusively on pixel-based RGB loss functions in this work:
1)~{\it L1} is a common choice, treating errors equally.
2)~{\it L2} penalizes larger errors more severely than smaller ones.
3)~{\it Log L1}, as its name suggests, is a log version of the L1 loss that tries to smooth the convergence curve, especially for large errors.
4)~{\it Relative L2} is more sensitive to cases where target pixels with high intensity are misaligned with less intense ones.
5)~{\it MAPE}~\cite{myttenaere2015esann:mape}, or ``Mean absolute percentage error'', is based on the relative percentage of errors. As the L1 equivalent of the ``Relative L2'' loss, it is scale-independent and places heavier penalty on negative errors.
6)~{\it sMAPE}~\cite{hyndman2006another}, the symmetric version of MAPE, may be unstable when both the prediction and ground truth have low intensity.
7)~{\it Smooth L1}~\cite{girshick2015fast} is designed to be less sensitive to outliers and can prevent exploding gradients (we set $\beta=0.1$ empirically). 

\subsection{Implementation Details}
In our experiments, the optimization process takes a total of $2560$ steps ($S=4$, so $s_1+Ss_2=2560$), where $P_N=64$, $s_1=s_2=512$ for the parallelized Monte Carlo sampling process.  We set $r=0.25$, which is halved each resampling round. We use the Adam optimizer~\cite{kingma2014adam} with learning rates that begin at $3 \times 10^{-3}$ for the translation part and $5 \times 10^{-3}$ for the rotation part, respectively. The learning rates decay exponentially with the base rate as 0.33 and base step as 256 over the course of optimization.
The whole process takes between 15 to 20~s depending on the size of the target on a single NVIDIA RTX 3090 GPU.

\section{Experimental Results \label{ExpResults}}

In this section, we demonstrate that our proposed method achieves improved robustness for both synthetic dataset and real-world scene compared to its predecessor iNeRF~\cite{yen2021inerf}.
We also explore the impact of using different pixel-based RGB losses for the optimization process. These results encourage further investigation to better model the difference between the registered target and the observed view.

\subsection{Synthetic Dataset \label{ExpResults_syn}}

{\bf Setting.} 
NeRF synthetic dataset~\cite{mildenhall2020nerf} consists of 8 geometrically complex objects with no background, including Chair, Drums, Ficus, Hotdog, Lego, Materials, Mic, and Ship. The objects have been resized to the unit box and are of complex non-Lambertian materials. Two of them (Ficus and Materials) are rendered from viewpoints sampled on a full sphere while the remaining six are rendered from viewpoints sampled on the upper hemisphere. The dataset provides camera intrinsics and extrinsics for each rendering as well as official splits for training/validation/test. All methods are trained on the training split views and evaluated on the test split for novel view pose estimation.

For each scene, we randomly choose 5 images from the test split and generate 5 different camera pose initializations. The starting pose is initialized by perturbing the ground truth pose:
we rotate the camera pose around its three axes sequentially by uniformly sampling from [-15, 15] degrees, then we translate the camera along the world axes by a random offset within [-0.25, 0.25] units.
 
We compare our proposed method with the state-of-the-art approach iNeRF~\cite{yen2021inerf}, where ``single'' and ``multiple'' denote our proposed without or with parallelized Monte Carlo sampling strategy, respectively. For a fair comparison, we use L2 loss for both of our method and iNeRF. For better accessibility, we use the vanilla NeRF model~\cite{mildenhall2020nerf} as iNeRF's basis. We re-trained the NeRF model~\cite{lin2020nerfpytorch} for 200k iterations with 64 samples for its coarse network and 128 samples for the fine network while setting the batch size to 4096. At inference time, we optimized iNeRF for 500 steps following its interest resampling strategy. The sampling number and batch size are set the same as training time.   In this way, we can make full use of the computing capacity and maximize its performance as noted by iNeRF~\cite{yen2021inerf}. The optimization process takes around 145~sec. for each object.

\begin{figure}[t]
  \centering
  \vspace*{0.06in}
  \begin{tikzpicture}[inner sep = 0pt, outer sep = 0pt]
    \node[anchor=south west] (fnC) at (0in,0in)
      {\includegraphics[width=\columnwidth,clip=true,trim=0in 0in 0in 0in]{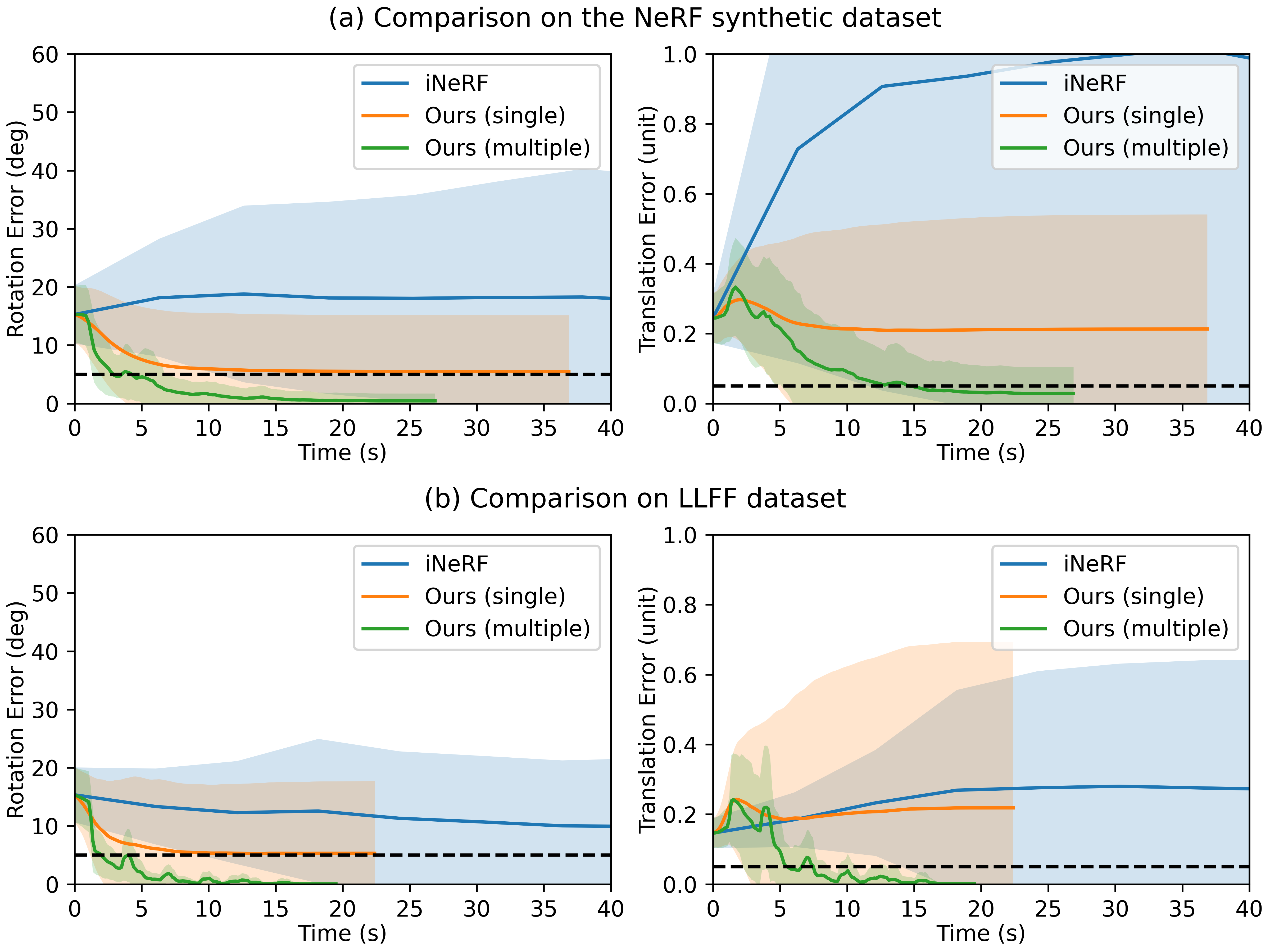}};
  \end{tikzpicture}
  \vspace{-0.3in}
  \caption{Quantitative comparison of the optimization time between iNeRF~\cite{yen2021inerf} and our proposed method on all the scenes from both synthetic dataset~\cite{mildenhall2020nerf} (top) and real-world dataset~\cite{mildenhall2019local} (bottom). Shown are the mean (solid lines), $\pm$1 standard deviation (shaded areas), and the criterion (5 degrees and 0.05 units) used in our experiment (black dotted line).
  As shown in the plot, iNeRF is vulnerable to large start pose error, and the update oftentimes drives it away from the target. Our single hypothesis variant performs better but gets trapped in local minima. 
  On the other hand, our proposed method with the parallelized Monte Carlo sampling module (the best camera pose hypothesis is shown) achieves more robust performance and converges to the target region faster.
  \label{fig:res_comp_quantitative}} 
  \vspace*{-15 px}
\end{figure}

\begin{figure}[t]
  \centering
  \vspace*{0.06in}
  \begin{tikzpicture}[inner sep = 0pt, outer sep = 0pt]
    \node[anchor=south west] (fnC) at (0in,0in)
      {\includegraphics[width=0.9\columnwidth,clip=true,trim=0in 0in 0in 0in]{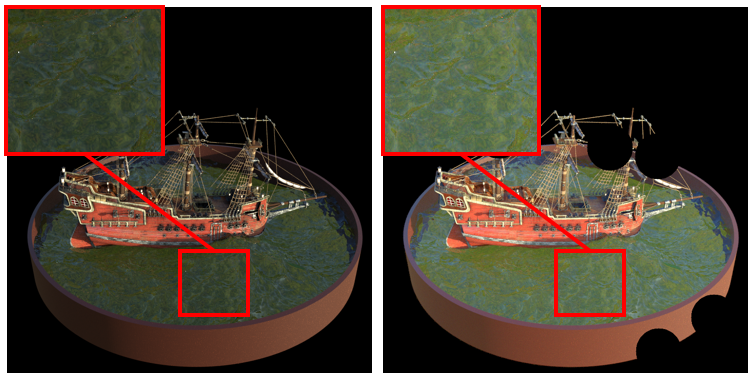}};
  \end{tikzpicture}
  \vspace{-0.1in}
  \caption{Visualization of the observed view from NeRF synthetic dataset~\cite{mildenhall2020nerf}. {\sc Left:} the original test image; {\sc Right:} the corrupted image with simulated Gaussian \& Poisson noise, brightness change, and the missing pixels. Note that the cropped region in the red bounding box (water) is textureless, making it hard to deal with.
  \label{fig:res_noisy}} 
  \vspace{-15px}
\end{figure}

{\bf Results.} 
We report the percentage of predicted poses whose error is less than 5 degrees or 0.05 units following~\cite{hodan2018bop}.
Table~\ref{tab:synthetic_and_llff}a highlights the performance of our proposed method. We make substantial improvements over iNeRF~\cite{yen2021inerf} on all the objects.
An example is shown in Figure~\ref{fig:res_comp_quantitative}a where the parallel Monte Carlo sampling strategy achieves better and faster evolution compared with the single pose hypothesis updates.
Noticeably, some targets, e.g., Mic and Ship, are more difficult. 
The thin body of the Mic is challenging for the optimization when the initial rendering has small overlap with the observed view, causing many samples to lie outside the object. 
Similarly, the large textureless areas (water) of Ship (see Fig.~\ref{fig:res_noisy}) cause sampling rays to receive a small loss when the rendered and the observed views are misaligned, an issue that becomes more severe as the number of sampling rays per hypothesis decreases under the same computation budget and fewer rays are aimed at the textured regions. We leave these challenges for future research on better importance sampling.

\begin{table*}[h]
 \vspace*{0.06in}
\caption{Ablation study on different pixel-based RGB losses. 
\label{tab:ablation}}
  \centering
\begin{tabular}{cccccccc}
\toprule

Dataset                & L1   & L2   & Log L1 & Relative L2 & MAPE & sMAPE & Smooth L1 \\
\midrule

\multicolumn{8}{c}{Rotation error $<5^{\circ}$ ($\uparrow$)} \\
\addlinespace

Synthetic~\cite{mildenhall2020nerf} (w/ simulated noise)  & 0.76 & 0.71 & 0.79  & {\bf 0.86}       & 0.82 & 0.70  & 0.77  \\
Real~\cite{mildenhall2019local} (reconstruction error) &{\bf  0.89} & 0.84 & 0.88  & 0.53       & 0.88 & 0.87  & 0.87  \\
Mean                  & 0.83 & 0.78 & 0.84  & 0.70       & {\bf 0.85} & 0.79  & 0.82 \\
\midrule
\multicolumn{8}{c}{Translation error $<0.05$ units ($\uparrow$)} \\
\addlinespace
Synthetic~\cite{mildenhall2020nerf} (w/ simulated noise)  & 0.65 & 0.58 & 0.70  & {\bf 0.78}      & 0.76 & 0.62  & 0.65   \\
Real~\cite{mildenhall2019local} (reconstruction error) & {\bf 0.87} & 0.78 & 0.86  & 0.37       & 0.82 & 0.85  & 0.85  \\
Mean                  & 0.76 & 0.68 & 0.78  & 0.57       & {\bf 0.79} & 0.73  & 0.75 \\

\bottomrule
\end{tabular}
\vspace{-15px}
\end{table*}

\subsection{Real-world Scene \label{ExpResults_real}}
{\bf Setting.} 
LLFF dataset~\cite{mildenhall2019local} is of complex real-world scenes captured with a handheld cellphone in a roughly forward-facing manner. 
Different scenes have images ranging from 20 to 62, while one-eighth of them is held out for the test split. We compare our method against iNeRF~\cite{yen2021inerf} on the four selected scenes: Fern, Fortress, Horns, and Room. To speed up the training process of iNeRF's basic NeRF model~\cite{lin2020nerfpytorch}, the high resolution images are downsampled by a factor of eight before feeding into all evaluated methods.

We adopt a similar procedure to generate start poses as described in Section~\ref{ExpResults_syn}. Considering that the views far away from the original camera center have too much artifact as all cameras are forward-facing, we change the translation perturbation range to [-0.15, 0.15] units following~\cite{yen2021inerf}.

{\bf Results.} We use the same metric in Section~\ref{ExpResults_syn} to measure performance. 
The results in Table~\ref{tab:synthetic_and_llff}b demonstrate that our proposed method with a single camera hypothesis has already improved over iNeRF~\cite{yen2021inerf} on all the real-world scenes. 
It indicates that our revised gradient policy (decoupled with angular momentum) based on Adam optimizer\cite{kingma2014adam} is helpful on the optimization process.
The parallelized Monte Carlo sampling strategy makes it even better as it can help alleviate the issue of trapping into local minima for a single camera pose hypothesis.

\subsection{Ablation Study on Different Pixel-based RGB Losses \label{ExpResults_ablation}}

{\bf Setting.} In this experiment, we are interested in the robustness of different pixel-based RGB losses to various errors introduced in the procedure. 
Since the NeRF synthetic dataset is rendered under a perfect simulation scenario, in addition to the procedure in Section~\ref{ExpResults_syn}, we simulate different kinds of disturbances on the test split images. They include environmental noise (Gaussian \& Poisson), lighting condition difference (brightness change), and missing pixels due to potential occlusion. A sample is shown in Figure~\ref{fig:res_noisy}.
The goal is to demonstrate the ability of the loss to handle potential visual difference between the rendered model (trained on the perfect simulation images) and the corrupted observed target image. Similar to Section~\ref{ExpResults_real}, we also evaluate different variants on the LLFF dataset~\cite{mildenhall2019local}. As the training images and observed test image are captured in the same sequence of a specific scene, the visual difference mainly comes from the reconstruction process. We compare seven variants with different pixel-based RGB losses described in Section~\ref{Approach_loss}.

{\bf Results.} As shown by the results in Table~\ref{tab:ablation}, our proposed optimization method differ significantly depending on the loss function used. In our context, the common practice of L2 loss~\cite{mildenhall2020nerf, yen2021inerf} does not perform as well as other loss options. Relative L2 loss performs best on the synthetic dataset with simulated noise while it is sensitive to the reconstruction error introduced in the real dataset. On the other hand, L1 gets the best results on the real dataset but slightly worse on the synthetic dataset. Overall, MAPE achieves the best balance across two datasets, serving as a better option to deal with various errors.  This finding echos the recent exploration of RawNeRF~\cite{mildenhall2022nerf} to handle high dynamic range scenes with a specially designed loss function.

\section{Conclusion}

We have proposed a parallelized optimization method based on Neural Radiance Fields (NeRF) for estimating 6-DoF poses with monocular RGB-only input. 
Our method performs accurate pose estimation with momentum-based camera extrinsics optimization integrated into a fast NeRF method.
We have demonstrated the advantage of parallelized Monte Carlo sampling in handling local minima, and its improved efficiency in a vast search space.
We have also shown the importance of a more robust pixel-based loss function for various errors.
The proposed method achieves improved robustness over both synthetic and real-world datasets. 
While our proposed method is currently optimized for offline applications, future research will be aimed at improving speed. Additionally, a better model of the visual difference (to handle, e.g., lighting differences and occlusion) between the registered model and the observed target is needed for enhanced robustness.

\section*{Acknowledgments}

We thank Chen-Hsuan Lin and Yen-Chen Lin for discussions related to the work, and 
Rogelio Olguin for assistance with the graphics. This work was supported in part by NSF Award \#2026611.

\bibliographystyle{IEEEtran}
\bibliography{main.bib}

\end{document}